\documentclass[10pt,journal, onecolumn]{article}

\usepackage{hyperref}
\usepackage{amsmath,amsfonts,amssymb,amsthm,mathtools}
\usepackage{graphicx,booktabs}
\usepackage{cleveref}
\usepackage{cite}
\usepackage{color}
\usepackage{csquotes}
\usepackage{mathtools}
\usepackage[margin=1in]{geometry}
\usepackage{array}
\usepackage{cases}
\usepackage{algorithm}
\usepackage{algpseudocode}
\usepackage{tikz}
\usetikzlibrary{shapes}
\usepackage{caption}
\usepackage{subcaption}

\usepackage{multirow}
\usepackage[table]{xcolor} 
\usepackage{colortbl} 
\usepackage{pifont}
\newcommand{\highlight}[2]{\colorbox{#1!20}{\textbf{#2}}}

\begin{document}

\title{Equivariant Learning for Unsupervised Image Dehazing}

\author{Zhang Wen\thanks{equal contribution, $\dagger$ corresponding author (d.chen@hw.ac.uk)}, Jiangwei Xie$^{*}$, Dongdong~Chen$^{\dagger}$ \\ Heriot-Watt University, Edinburgh, UK
}
\date{}
\maketitle

\begin{abstract}
Image Dehazing (ID) aims to produce a clear image from an observation contaminated by haze. Current ID methods typically rely on carefully crafted priors or extensive haze-free ground truth, both of which are expensive or impractical to acquire, particularly in the context of scientific imaging. We propose a new unsupervised learning framework called Equivariant Image Dehazing (EID) that exploits the symmetry of image signals to restore clarity to hazy observations. By enforcing haze consistency and systematic equivariance, EID can recover clear patterns directly from raw, hazy images. Additionally, we propose an adversarial learning strategy to model unknown haze physics and facilitate EID learning.  Experiments on two scientific image dehazing benchmarks (including cell microscopy and medical endoscopy) and on natural image dehazing have demonstrated that EID significantly outperforms state-of-the-art approaches.  By unifying equivariant learning with modeling haze physics, we hope that EID will enable more versatile and effective haze removal in scientific imaging. Code and datasets will be published.
\end{abstract}

\section{Introduction}\label{sec:intro}

Image Dehazing (ID) is a fundamental inverse problem in computer vision, computational imaging, and signal processing. It aims to restore the clarity and contrast of images degraded by smoke or atmospheric haze. ID is essential for enhancing visual quality in a wide range of applications, including autonomous driving, surveillance, healthcare diagnostics, remote sensing, and biomedical imaging.


\begin{figure}[!htbp]
    \centering
    \begin{subfigure}{\columnwidth}
        \centering
        \includegraphics[width=0.95\linewidth]{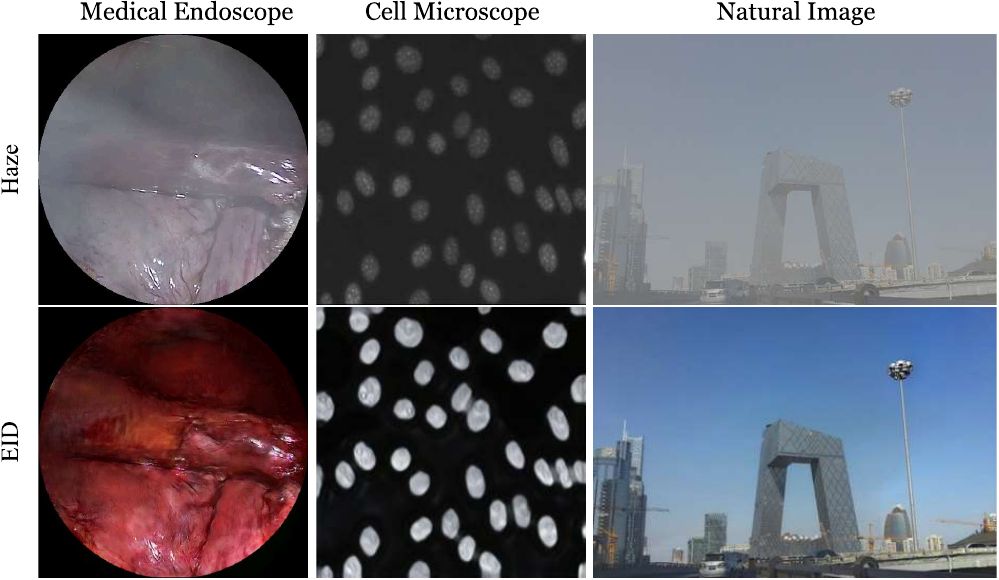}
        \caption{\scriptsize Examples of dehazing performance for different applications. Top are collected haze observations, while the bottom are dehazed images by EID.}
    \end{subfigure}
    \vskip\baselineskip
    \begin{subfigure}{\columnwidth}
        \centering
        \includegraphics[width=0.9\linewidth]{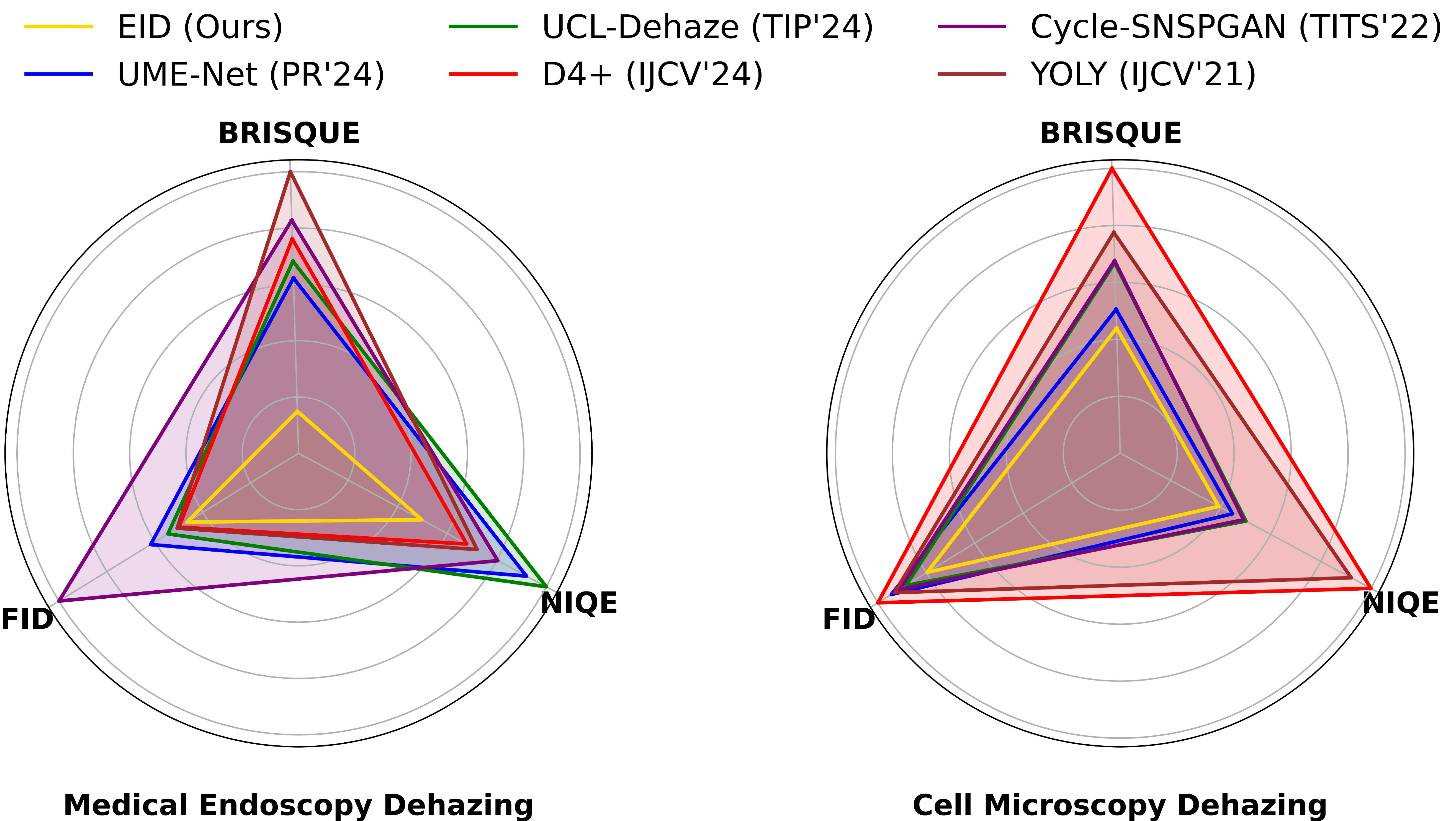}
        \caption{\scriptsize EID achieves the state-of-the-art performance on Cholec80-Haze \cite{cholec80data} (medical endoscopy) and Cell97 \cite{kagglecell} (cell microscopy) in 3 metrics: NIQE$\downarrow$, BRISQUE$\downarrow$, FID$\downarrow$ metrics are calculated, with lower values indicating better performance.}
    \end{subfigure}
    \caption{Performance comparison of our proposed EID with existing image dehazing approaches for scientific imaging and natural image restoration.}
    \label{fig:radar}
\end{figure}

Generally, ID involves inverting the haze degradation process described by:
\begin{equation}
    y = \mathcal{H}(x)
\end{equation}
where \( y \) is the observed hazy image, \( x \) is the underlying clear image, and \( \mathcal{H} \) represents the nonlinear haze degradation process. In principle, solving the ID problem entails maximizing the log-likelihood $\log p(x|y)$ which is equivalent to solve the below optimization: 
\begin{equation}
    \min_x \{\ell(\mathcal{H}(x),y) + \mathcal{R}(x)\}
\end{equation}
where $\ell$ is for haze consistency and $\mathcal{R}$ is the regularization term.

In the literature, ID methods can be categorized into two primary groups: \emph{prior-based} \cite{ur2p,pr2025novel} and \emph{learning-based} approaches \cite{pr2023deep}. Prior-based methods use domain knowledge to design hand-crafted priors $p(x)$ that facilitate plausible restoration of hazy images, relying on prior knowledge of clean images and the physics of the haze process. In contrast, the learning-based approaches aim to formulate a data-driven prior $x$ by learning direct mappings between hazy images \( \{y\} \) and their corresponding ground truth clear images \( \{x\} \), which achieve state-of-the-art dehazing performance \cite{odcr-cvpr-2024-intro} on natural image restoration. 
However, in scientific image dehazing applications such as cell microscopy, remote sensing and medical endoscopy (e.g. water vapour and smoke generally obscure the scene), obtaining ground truth is either very expensive or even impossible. This raises a critical question: \emph{can we avoid relying on strong image priors or on acquiring ground truth data for ID?}

In this paper, we present a novel, fully unsupervised framework: \emph{Equivariant Image Dehazing} (\textbf{EID}) for learning dehazing from raw haze images. Our approach is based on the observation that the set of clear images generally possesses symmetry \cite{spm}, i.e. invariant to certain transformation groups, therefore the whole haze-dehaze process is equivariant. By enforcing haze consistency and systematic equivariance, EID enables neural network dehazing from only raw haze images. The EID framework is agnostic to neural network architectures, allowing integration with existing dehazing networks or  handcrafted blocks. 
In addition to the EID framework, we also present cyclic adversarial training to model the haze process $\mathcal{H}$ for cases that are unknown.
Extensive experimental results demonstrate EID outperforms existing state-of-the-art approaches on scientific image dehazing tasks including cell microscopy and medical endoscopy, and natural image dehazing. We hope that this approach will provide a promising alternative to image dehazing, particularly in scientific imaging scenarios, see Figure~\ref{fig:radar}. In summary, our key contributions are as follows:
\begin{itemize}
    \item We present EID, a novel \emph{fully unsupervised} and physics-informed image dehazing paradigm that leverages symmetry (invariance) of natural images. 
    \item We present an end-to-end, adversarial learning strategy for modeling unknown haze physics and facilitating EID learning.
    \item Extensive experiments demonstrate that EID achieves state-of-the-art dehazing performance in \emph{medical endoscopy}, \emph{cell microscopy} and \emph{natural image} restorations.   
\end{itemize}

\noindent\textbf{Notations.} In this paper, given a matrix $A\in\mathbf{R}^{m\times n}$, $A^{\dagger}$ is the pseudo-inverse of $A$, we denote the nullspace of $A$ as $\mathcal{N}_A=\{x \mid Ax=0, \forall x\in\mathbf{R}^n\}$ and the range space of $A$ as $\mathcal{R}_A= \{Ax \mid \forall \ x\in \mathbf{R}^n\}$. According to the range-nullspace decomposition of signals \cite{ddn}, we have $\mathcal{R}_{A^{\dagger}}\oplus\mathcal{N}_A=\mathcal{R}^n$.

\section{Related Work}
\label{sec:relateWork}

\subsection{Image Dehazing}
Image dehazing (ID) has been a long-standing research topic in computer vision, aiming to restore the visibility of images degraded by atmospheric particles such as fog or haze.
Early classical methods are predominantly based on the atmospheric scattering model~\cite{scatter-1975-intro}, which describes the physical process of light transmission in hazy environments and enables principled image restoration via handcrafted priors.
A seminal contribution is the Dark Channel Prior (DCP) proposed by He et al.~\cite{he-DCP-2010-intro}, which leverages statistical properties of outdoor images to estimate scene transmission maps with remarkable effectiveness.
Following this line, numerous priors have been introduced, including color-line prior~\cite{fattal-clp-prior2014-intro} and non-local prior~\cite{berman-ncp-prior2016-intro}, each exploiting different statistical regularities for improved robustness.

Despite their success under certain conditions, these traditional approaches often struggle when confronted with complex scenes where underlying assumptions about image statistics or physical parameters may not hold universally.
To overcome reliance on strong but potentially brittle priors, recent years have witnessed a paradigm shift towards data-driven techniques powered by deep learning.

Supervised deep learning-based ID methods typically learn an end-to-end mapping from paired hazy inputs to corresponding clean ground-truth images~\cite{convir-2024pami-surv-baseline,luo2023diffusion-relate}. For instance, Li et al.~\cite{aodnet-relate} designed an all-in-one dehazing network that enables direct regression from input to output without explicit estimation of intermediate variables; Guo et al.~\cite{guo2022ideharmber-relate} utilized hybrid CNN-transformer architectures that aggregate local details and global context simultaneously; Luo et al.~\cite{luo2023diffusion-relate} adopted latent diffusion models, achieving superior performance through generative modeling in high-dimensional latent spaces.

Nevertheless, supervised approaches require large-scale paired datasets for training—resources that are costly or infeasible to obtain especially for real-world scenarios involving diverse weather conditions or sensor modalities. Motivated by advances in unsupervised domain adaptation and adversarial learning frameworks such as CycleGAN~\cite{cycleGAN-intro}, recent studies explore unpaired image-to-image translation paradigms for ID tasks. CycleDehaze~\cite{cycledehze-intro} exploits cyclic perceptual consistency loss to distill haze removal knowledge from unpaired samples;
UCL-DeHaze~\cite {ucldehaze-intro} combines patch-wise constraints with contrastive losses tailored specifically for dehazing applications; D4 framework explicitly incorporates physics-based scattering coefficients and depth cues into its generation pipeline so as to synthesize variable-density hazy imagery during training~\cite{d4-relate}; its successor D4+ further introduces dual contrastive perceptual objectives enhancing restoration quality~\cite{d4+-baseline}. UMENet \cite{umenet-2024-pr-baseline} addresses distribution confusion issues via compensation strategies targeting high-frequency information recovery.

Although significant progress has been made across both supervised and unsupervised domains—with increasingly sophisticated architectures integrating multi-level features—the majority of existing works remain dependent either on stringent hand-crafted priors or on access to ground truth reference data. This dependency restricts their general applicability when applied to challenging real-world and scientific imaging situations where labeled pairs are unavailable or where scene characteristics deviate substantially from those assumed during development. In this paper, we propose a fully new paradigm that is completely unsupervised for solving real world image and scientific imaging dehazing problems, see Figure~\ref{overview}.

\subsection{Equivariant Imaging}
Equivariant Imaging (EI) \cite{dongdongEI2021-intro,REIdongdong-relate} is a state-of-the-art fully unsupervised framework for solving inverse imaging problems without accessing ground truth, and has been successfully applied to various computational imaging and low-level computer vision tasks \cite{EMMA,pantylirotate-2024arxiv-transform}. 
In particular, EI leverages the group invariance properties inherent in natural signals to learn a reconstruction function using only partial measurement data. It operates under the assumption that the original signal set possesses symmetry properties and remains invariant under a group of transformations, such as rotation, shifting, and scaling. 

Let $\mathcal{X}$ denote the set of ground truth images, and let $\{T_g\}$ denote the group of actions (e.g. rotations). If, for all $\forall x\in \mathcal{X}, T_gx \in \mathcal{X}$, then $\mathcal{X}$ is invariant under $T_g$. If a function satisfies the condition that $T_g\mathcal{F}(x) = \mathcal{F}(T_gx)$, then it is equivariant \cite{dongdongEI2021-intro,EMMA} and the restoration function $f: y\rightarrow x$ can be learnt by enforcing equivariance across the entire imaging system, i.e. $\mathcal{F}=f\circ A$, without accessing the ground truth image set \cite{dongdongEI2021-intro,dongdongJMLRsensing-relate}. As a pioneering imaging approach, EI shows great promise in solving many linear inverse problems. This paper aims to explore the potential of EI in tackling a more complex challenge: nonlinear and blind inverse problems in image dehazing.

\section{Method}
\label{Motivation}
\begin{figure*}
    \centering
    \includegraphics[width=\linewidth]{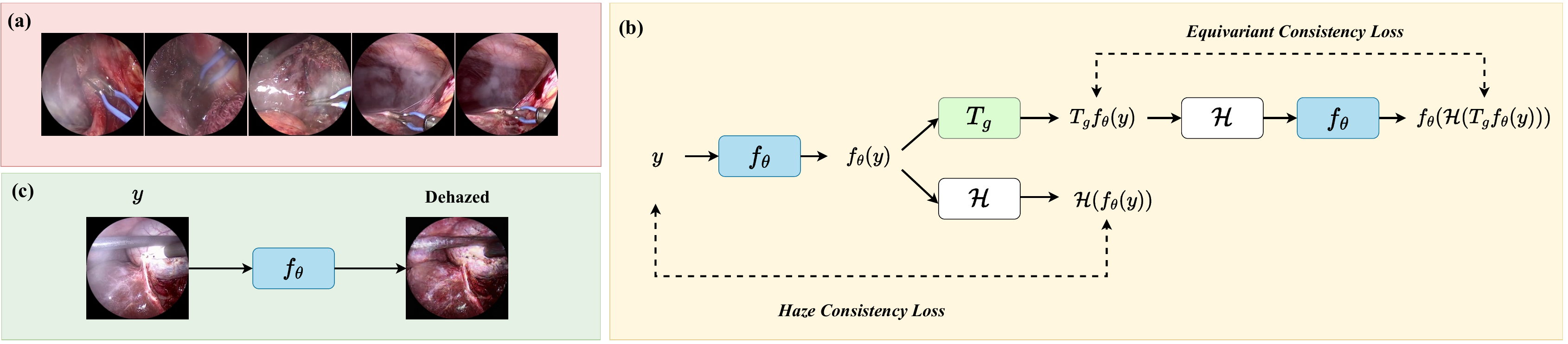}
    \caption{Overview of EID. (a) \textbf{Motivation}: scientific or natural images often show invariances to certain transformations (e.g. rotation), as seen in the real endoscopic examples above. (b) \textbf{Training without ground truth}: given haze images \(\{y\}\), the dehazing model \(f_\theta\) first outputs coarse dehazed images \(f_\theta(y)\), which are then fed back into the haze model to minimize haze consistency. At the same time, \(f_\theta(y)\) is transformed by group action \(T_g\) and perform haze and dehaze again to achieve system equivariance. By continuing the training loop, \(f_\theta\) learns the fine and clean dehazed image. (c) \textbf{Testing}: the EID-trained \(f_\theta\) can be used directly to dehaze new, unseen haze images.}
    \label{overview}
\end{figure*}

In scientific imaging scenarios, obtaining ground truth images in real-world scenarios is often impractical. For example, in endoscopic imaging \cite{endoscope-imaging-survey-2014}, haze caused by steam inside the body can obscure details, making it difficult to obtain clear, haze-free images. Consequently, in ground truth-free settings, the training set will only contain haze images \(\{y\}\), and the learning objective must then be reformulated to avoid reliance on ground truth data. In the following, we will show that this can be achieved by learning with physics $\mathcal{H}$-informed haze consistency and equivariance constraints.

\subsection{Haze Consistency} In general, the optimal dehazing function $f_\theta:y\rightarrow x$ should maintain consistency within the haze image domain governed by the haze process $\mathcal{H}$:
\begin{equation}
\mathcal{H}(f_{\theta}(y)) = y
\end{equation} 

However, due to the underdetermined nature of the haze process, the estimation of $f_{\theta}$ cannot be achieved by minimizing the error between $\mathcal{H}(f_{\theta}(y))$ and $y$ for two reasons. Firstly, we don't have the nullspace information of $\mathcal{H}$ w.r.t. the ground truth, i.e. the clear components in the ground truth image, but only access the information in its range space, i.e. the haze components in $y$, and we need to learn more information beyond these range space components \cite{dongdongEI2021-intro}. Secondly, the physics of $\mathcal{H}$ is generally inaccessible. To address these challenges, we first propose a pseudo-sensing module to model the physics of the haze process $\mathcal{H}$, and then introduce an equivariant dehaze module to learn clear image components such that the groud truth $\{x\}$ can be recovered from its raw haze images $\{y\}$.

\subsection{Pseudo Hazing Module}\label{sec:pseudo_h}
Ideally, the haze process $\mathcal{H}$ can be approximated by a conventional scattering model \cite{scatter-1975-intro}:
\begin{equation}
\label{haze physics}
    \mathcal{H}(x) \approx x \cdot exp(-\beta d) \;+\; \alpha\,\bigl(1 - exp(-\beta d)\bigr)
\end{equation} where $x$ is the clean image, $\beta$ is the attenuation coefficient, $d$ is the depth, and $\alpha$ is the atmospheric light parameter. In general, the hyperparameters of $\mathcal{H}$ in Eq.~\ref{haze physics} can be well estimated by e.g. \cite{d4+-baseline,ridcp} for natural image dehazing. 

\begin{figure}
    \centering
    \includegraphics[width=1\linewidth]{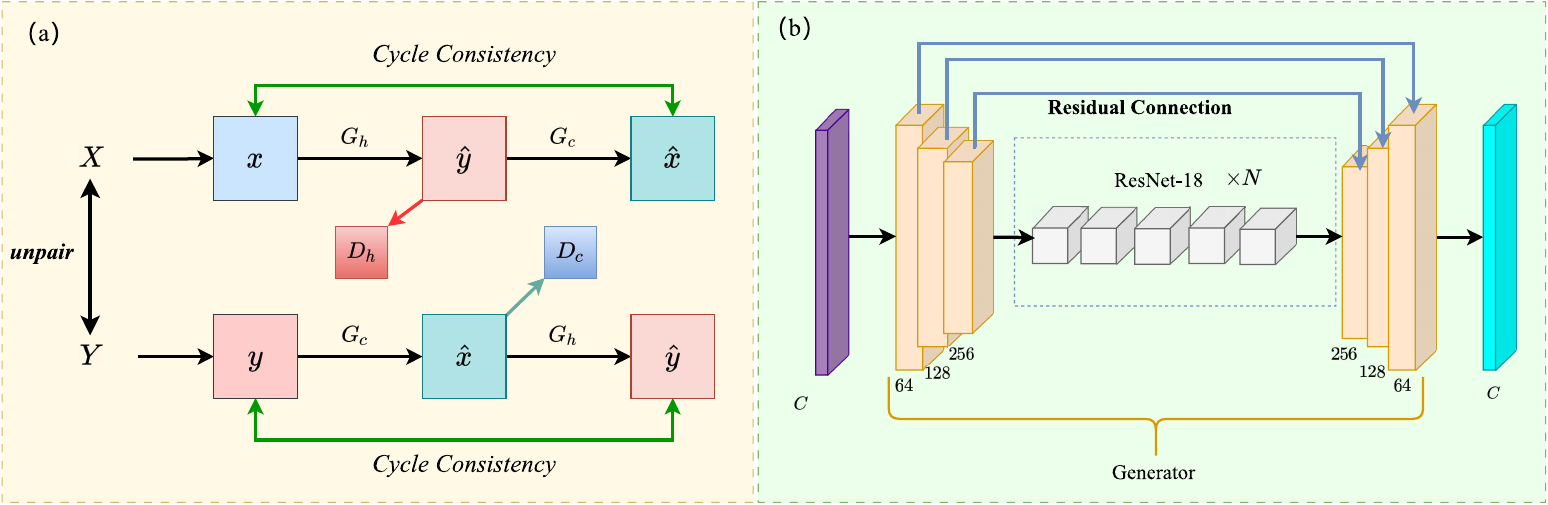}
    \caption{(a) Adversarial training of the generator $G_h$ for modelling the hazing physics. (b) The proposed architecture of $G_h$.}
    \label{fig:GAN}
\end{figure}
However, the haze processes in scientific imaging applications such as endoscopy and cell microscopy are more complicated and not well characterised by simple analytical models in (\ref{haze physics}), requiring the learning of \emph{pseudo} physics $\mathcal{H}$. To this end, we apply cyclic adversarial learning on unpaired haze and clear images \cite{cycleGAN-intro} to effectively model the physics of $\mathcal{H}$ as well as the distribution of haze images $p(y)$. 
In particular, to effectively model the mapping between clear and hazy image domains, we train a generator $G_h$ using adversarial training on unpaired data; see Figure~\ref{fig:GAN}. This ensures that the generated hazy images resemble real-world haze patterns and maintain structural consistency when converted back to the clear domain.

Firstly, to generate realistic hazy images, we propose training a generator $G_h$ for converting clear images into hazy ones using the adversarial loss pioneered by \cite{gan-valinna-sm}, i.e.
\begin{equation}\label{eqs:gan_goodfellow}
\ell_1=\mathbb{E}_{x_{h}}\left[\log D_{h}(x_{h})\right] +\mathbb{E}_{x_{c}}\left[\log \left(1-D_{h}(G_{h}(x_{c}))\right)\right]
\end{equation}
where $x_c$ and $x_h$ denote samples from the distributions of clear and hazy images, respectively. The discriminator $D_h$ aims to distinguish between real hazy images and generated ones.  The aim of $G_h$ is to produce hazy images that are indistinguishable from real ones and thus effectively fool $D_h$. 

Although the adversarial training in Eq.(\ref{eqs:gan_goodfellow}) ensures that the generated images resemble real hazy samples, it does not guaranty structural consistency. To address this issue, we propose adopting a cycle consistency loss \cite{cycleGAN-intro}, which enforces the constraint that an image translated to the opposite domain and then mapped back should remain close to the original input:
\begin{equation}
\ell_2=\mathbb{E}_{x_{h}}\left[\|G_{h}(G_{c}(x_{h}))-x_{h}\|\right] + \mathbb{E}_{x_{c}}\left[\|G_{c}(G_{h}(x_c))-x_{c}\|\right]
\label{eq:cyc-loss}
\end{equation}
where $G_c$ translates hazy images back to the clear domain.

Finally, by minimizing the combined loss $\ell_1 + \ell_2$, the generator ($G_h$) that translates clear images into hazy ones ensures that the generated images resemble real hazy samples, while also guaranteeing structural consistency, i.e., no unnecessary distortions are introduced during the translation process. Through adversarial training, the model learns pseudo-haze physics that approximate real-world characteristics. Once trained, $G_h$ is therefore  used to estimate the real haze function (i.e. $G_h\to\mathcal{H}$) and is kept \emph{frozen} in the EID pipeline.

\textbf{Remark.} We emphasize the following: (1) One might argue that $G_c$ already constitutes a mapping that can convert blurred images into clear ones and be used directly for dehazing. However, subsequent experiments demonstrate that its efficacy cannot guaranty the accuracy of the resulting clear images, as the ID problem is fundamentally ill-posed. Therefore, to enable the machine dehazing systems to produce more plausible clear images, we must provide them with additional, relatively mild prior knowledge of clean images. (2) Our pseudo-hazing module $G_h$  is not specific to building a data-driven $\mathcal{H}$, and here we present an effective method of achieving this through adversarial learning. Alternatively, to model the haze function $\mathcal{H}$, one could first use state-of-the-art dehazing methods to generate a haze-reduced image, which can then be used as the golden-standard reference for the original hazy image. A direct mapping can then be trained from the reference image to the target hazy image, as in \cite{EMMA}.  (3) More importantly, even if $\mathcal{H}$ is given, the challenge of learning to dehaze from the raw haze image alone remains, which motivates our equivariant dehazing approach, as presented below. 

\subsection{Equivariant Consistency} 

In scientific image dehazing tasks, the object of interest is usually captured under different sensing conditions, resulting in images with domain-specific symmetry, i.e. invariance with respect to certain transformation groups \cite{spm,dongdongJMLRsensing-relate}. For example, in endoscopic imaging (Figure~\ref{overview}a), organs such as the stomach and intestine move continuously due to peristalsis and the endoscope's orientation changes, leading to clear rotation invariance. Similarly, although the microscope lens may remain stationary in cellular imaging, cells are often placed in different orientations on slides, leading to data that is invariant to translation, scaling and rotation. In principal, if the image set $\{x\}$ is invariant to a transformation group $\{T_g\}$, the whole haze-dehaze system  $f_\theta\circ \mathcal{H}$ should be equivariant to that transformations, i.e. 
\begin{equation}\label{eqs:equi}
    f_\theta \bigl(\mathcal{H}(T_{g} x)\bigr) = T_{g} \bigl( f_\theta(\mathcal{H}(x)) \bigr).
\end{equation}

Since the ground truth $\{x\}$ is not available, as inspired by \cite{dongdongEI2021-intro}, we use $f_\theta(y)$ as an estimate of $x$ on the left side of Eq.(\ref{eqs:equi}) and rewrite the right side as $T_{g}( f_\theta(y))$ by the fact $\mathcal{H}(x)=y$, which leads to the following ground truth-free equivariance constraint:
\begin{equation}
    f_\theta(\mathcal{H}\bigl(T_{g} f_{\theta}(y)\bigr)) = T_{g}f_{\theta}(y).
\end{equation}

\noindent\textbf{Equivariant Image Dehazing.} Finally, by integrating the haze consistency constraint and the equivariance constraint, the proposed fully unsupervised (ground truth free) EID loss is as follows:
\begin{equation}
\label{final_loss}
\arg\min_{\theta}\mathbb{E}_{y, g}\left\{\underbrace{\mathcal{L}\bigl(\mathcal{H}(f_\theta(y)), y\bigr)}_{\mathcal{L}_{hc}}+ \lambda\underbrace{ \mathcal{L}\bigl(f_\theta(\mathcal{H}(T_{g} f_{\theta}(y))), T_{g}f_{\theta}(y)\bigr)}_{\mathcal{L}_{ec}}\right\}
\end{equation} 
where $\mathcal{L}_{hc}$ and $\mathcal{L}_{ec}$ represent haze consistency and system equivariance, respectively. The parameter $\lambda$  balances the two constraints. $\mathcal{L}$ is the error function, e.g. mean squared error (MSE). The entire training pipeline is shown in Figure~\ref{overview} (b), and the trained haze removal network, $f_\theta$, can be used to remove haze from newly acquired hazy images, as shown in Figure~\ref{overview} (c).

\section{Experiments}

In this section, we aim to explore the following questions: 
\begin{enumerate}
    \item \emph{Can the proposed EID help the neural network achieve superior dehazing results in scientific imaging applications and natural image restoration?}
    \item \emph{How important are the hazy consistency ($\mathcal{L}_{hc}$) and equivariance consistency ($\mathcal{L}_{ec}$) components to EID? }
    \item \emph{What impact does using different transformation groups $\{T_g\}$ have?}
\end{enumerate}

We first introduce the experimental setup, followed by results on two scientific image dehazing scenarios and one natural image dehazing task, and conclude with ablation studies on loss functions and transformation groups.

\subsection{Experimental Setup} We conduct experiments on two benchmark scientific imaging tasks including endoscope (\emph{Cholec80-Haze} dataset \cite{cholec80data}) and cellular microscopy (\emph{Cell97 dataset} \cite{kagglecell}) where the ground truth clear images are not accessible, and also the natural image dehazing (\emph{RESIDE} dataset \cite{RESIDE-DATASET}).

\noindent\textbf{SOTA methods and metrics.} We compare our EID with nine state-of-the-art methods, including two prior-based approaches: DCP \cite{he-DCP-2010-intro} and NLP \cite{berman-ncp-prior2016-intro}, as well as seven deep learning-based unsupervised algorithms: CycleGAN \cite{cycleGAN-intro}, Cycle-Dehaze \cite{cycledehze-intro}, Cycle-SNSPGAN \cite{wang2022cycle-intro}, YOLY \cite{yoly-baseline}, UCL-Dehaze \cite{ucldehaze-intro}, D4+ \cite{d4+-baseline}, and UME-Net \cite{umenet-2024-pr-baseline}.
Five metrics are used to evaluate dahazed image quality, including NIQE, BRISQUE and FID for dehazing quality in scientific imaging, and PSNR and SSIM for natural images dehazing where ground truth labels are available.
\begin{figure}[!htbp]
    \centering
    \includegraphics[width=0.90\linewidth]{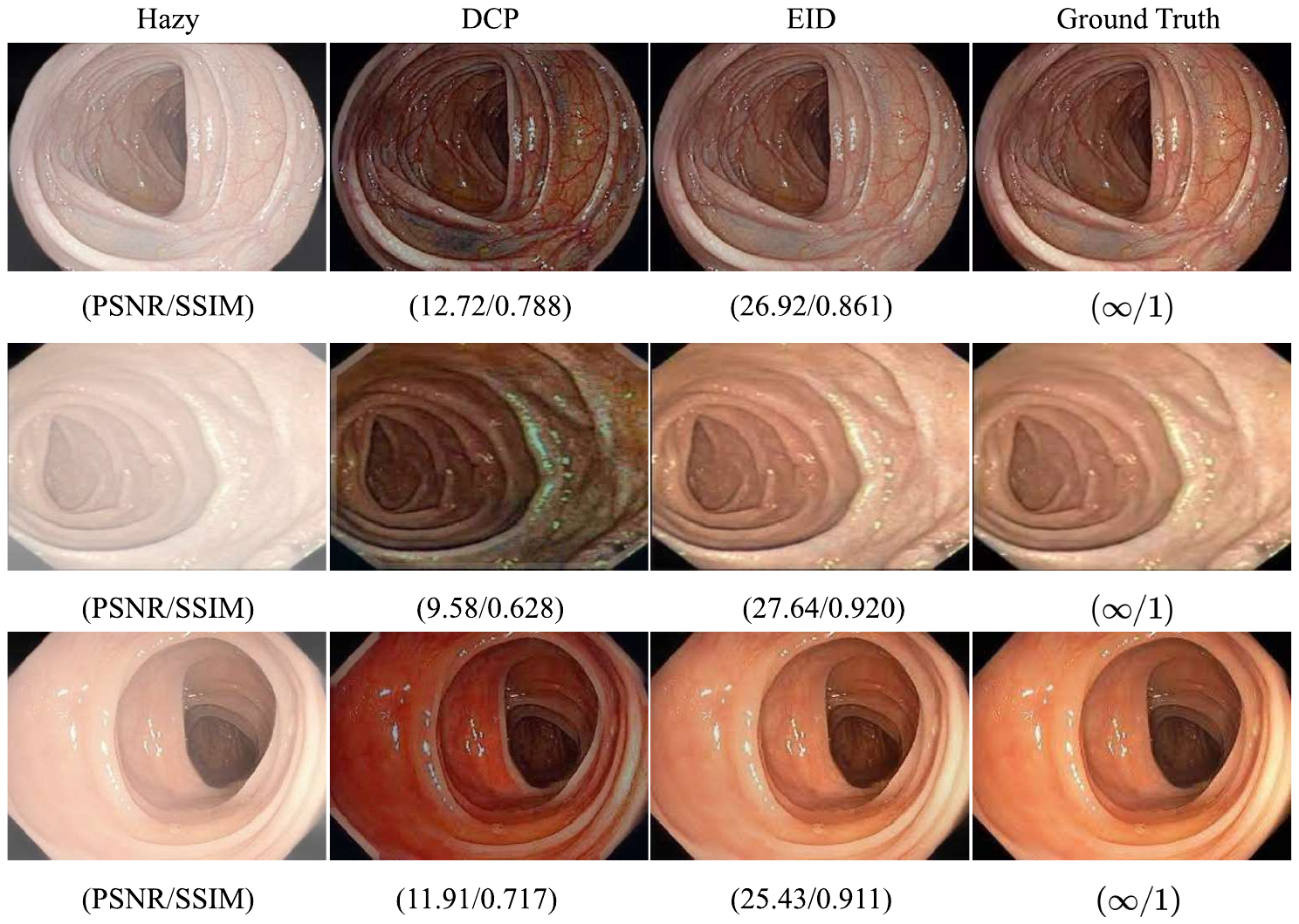}
    \caption{Comparison of endoscopy image dehazing performance between EID and DCP. We use the Dense Prediction Transformer \cite{dpt-depthpredict-sm} to predict the depth map and add haze following Eq.\ref{haze physics}. PSNR $\uparrow$ and SSIM $\uparrow$ values are displayed below each image.}
    \label{sm-medical-motivation}
\end{figure}

\begin{figure}
    \centering
    \includegraphics[width=0.95\columnwidth]{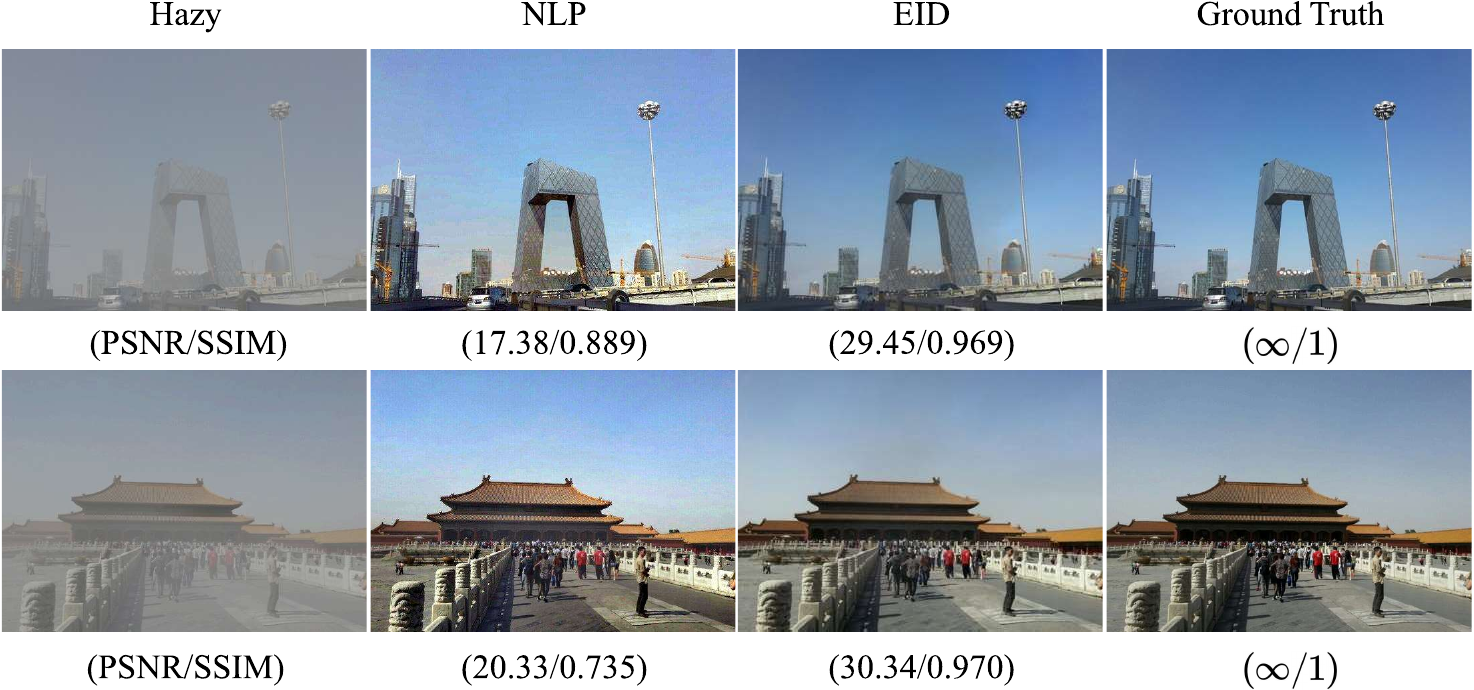}
   \caption{Comparison of natural image dehazing performance between EID and NLP \cite{berman-ncp-prior2016-intro}. We utilize provided depth map in RESIDE dataset and follow Eq.(\ref{haze physics}) to add haze. PSNR $\uparrow$ and SSIM $\uparrow$ values are shown below each image.}
    \label{sm-motivation-natural}
\end{figure}

\noindent\textbf{Implementation.} We implement all experiments based on PyTorch and the DeepInverse \cite{deepinv} library, using a single NVIDIA GeForce RTX 3090 GPU. We use rotation as the transformation group for EID. We use the Adam optimizer for 50 epochs, starting with a learning rate of $10^{-4}$ (halved every 20 epochs). We set $\lambda=0.1$ in Eq.~\ref{final_loss}. The mesh $f_{\theta}$ is a 5-layer U-net \cite{unet}. For the endoscope, cell microscope and natural image dehazing, whose $\mathcal{H}$ are unknown, we trained a UNet-like network as an estimated pseudo-physics to $\mathcal{H}$ using cyclic adversarial learning as presented in Sec.~\ref{Motivation}. We also test EID on the cases where $\mathcal{H}$ is known. We demonstrate that EID remains effective when the haze physics is well-defined and accurate, eliminating the need to learn a pseudo-physics model through adversarial learning, as discussed in Sec.\ref{Motivation}. Figure~\ref{sm-medical-motivation} presents dehazing results on medical endoscopy images, while Figure~\ref{sm-motivation-natural} shows results on natural images. EID outperforms prior-based methods in both tasks.

\subsection{Medical Endoscopy}


In medical endoscopic imaging, haze is commonly encountered \cite{endoscope-imaging-survey-2014}, but no prominent and public dataset has been established for this specific task. To this end, we collected a real-world dataset \emph{Cholec80-Haze} based on the public Cholec80 collection \cite{cholec80data}, which provides 80 videos of cholecystectomy procedures annotated with surgical phases and instrument presence. From these videos, we selected various clear and hazy frames, each at a resolution of $480\times480$, to curate the high-quality real-haze endoscopic dataset.
In particular, we first use 2726 clear images and 1100 unpaired hazy images to train the pseudo-haze module (i.e. $G_h\rightarrow \mathcal{H}$) using Sec.~\ref{sec:pseudo_h}. We then use the 1,100 hazy images as the training set to learn the EID dehazing model, and the remaining 146 hazy images as the test set.

\begin{figure}[!htbp]
    \centering
    \includegraphics[width=1\linewidth]{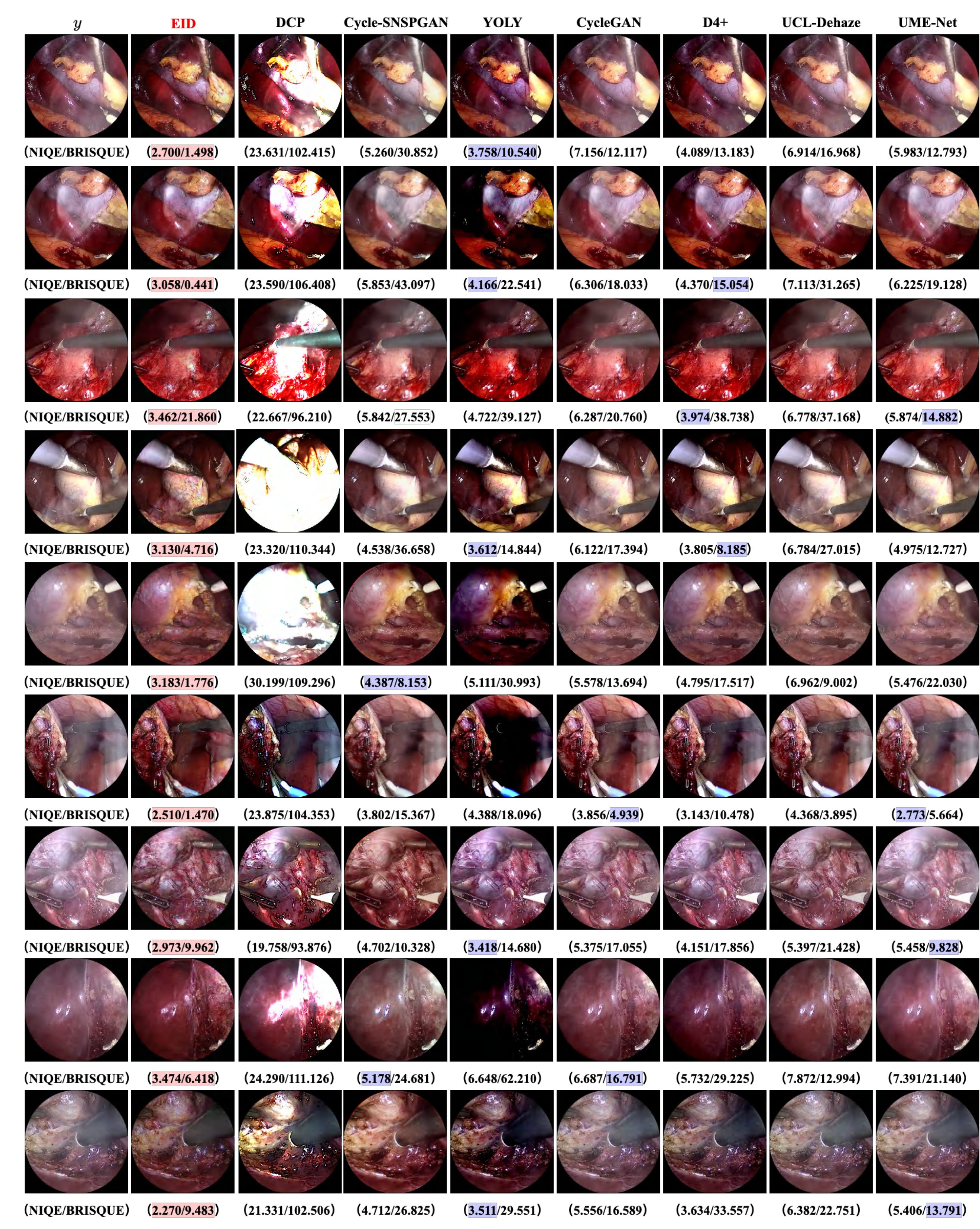}
    \caption{Results on Cholec80-Haze. Corresponding NIQE$\downarrow$ and BRISQUE$\downarrow$ values are presented under the pictures. The best is \highlight{red}{red}, and the second-best is \highlight{blue}{blue}.}
    \label{cholec80-result}
\end{figure}

The quantitative comparisons are presented in Figure~\ref{cholec80-result} and Tab.~\ref{quantitative}. We have the following observations: EID outperforms all its counterparts in terms of sharper appearance, more appropriate colors, richer fine details, and the lowest NIQE, BRISQUE and FID. DCP tends to suffer from overexposure. Cycle-SNSPGAN over-enhances certain details and alters the color tone. YOLY introduces visible artifacts and noise. CycleGAN, i.e. $G_c$ in Eq. (\ref{eq:cyc-loss}), over-smooths details. D4+ shows color distortion. UCL-Dehaze output is often too dark or blurred. UME-Net occasionally produces blurred or jagged edges.

\subsection{Cellular Microscopy} 
In the microimaging environment, cellular images are often degraded by scattered light, refractive attenuation of multiple layers of biological tissue, and lens stray light, necessitating the effective removal of such haze-like artifacts, while it's very expensive to acquire the ground truth clean images to learn from \cite{belthangady2019applications}.

We evaluate our method on the \emph{Cell97} dataset \cite{kagglecell}, which contains 97 either clear or haze-like (i.e. unpaired) fluorescence microscopy images and a total of 4009 cells. In particular, the low-quality images were acquired via an automated fluorescence microscope (Beckman Coulter IC-100 Company), while the high-quality images were captured using an Opera automated microscope equipped with Nipkow confocal spinning disks (Perkin-Elmer Company). Although “haze” here denotes poor contrast or blurring, it mirrors haze artifacts found in outdoor or endoscopic imaging. Consequently, true ground-truth images are unavailable. Finally, we use the 49 high quality and 48 unpaired low quality images as the training set and all of the low quality images as the test set.

\begin{figure}[!htbp]
    \centering
    \includegraphics[width=0.45\linewidth]{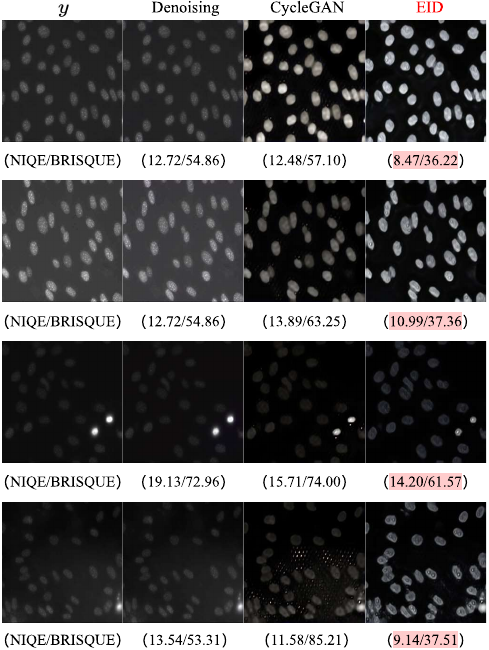}
    \caption{Performance comparison among denoising algorithms, CycleGAN and our proposed EID. The results suggest that cell microscopy imaging is better characterized as a dehazing problem rather than a denoising one. The \highlight{red}{red} marker represents the best value.}
    \label{denoising}
\end{figure}

Before evaluating, we investigate whether the cell microscopy task on the Cell97 dataset is better characterized as a \textit{haze removal} task rather than a \textit{denoising} task. To this end, we use UNSURE-Gaussian loss \cite{UNSURE-mikedavis-sm} to train a denoising network using a 5-layer U-Net \cite{unet}, optimizing with Adam for 100 epochs, starting with a learning rate of $1\times10^{-4}$, and other training settings follow the EID setup. To evaluate the distinction between dehazing and denoising, we compare the results from both approaches, as shown in Figure~\ref{denoising}. It can be observed that the dehazing results produce clearer images compared to the denoising outputs, further supporting our hypothesis that the task is better framed as dehazing rather than denoising.

\begin{figure}[!htbp]
    \centering
    \includegraphics[width=0.9\linewidth]{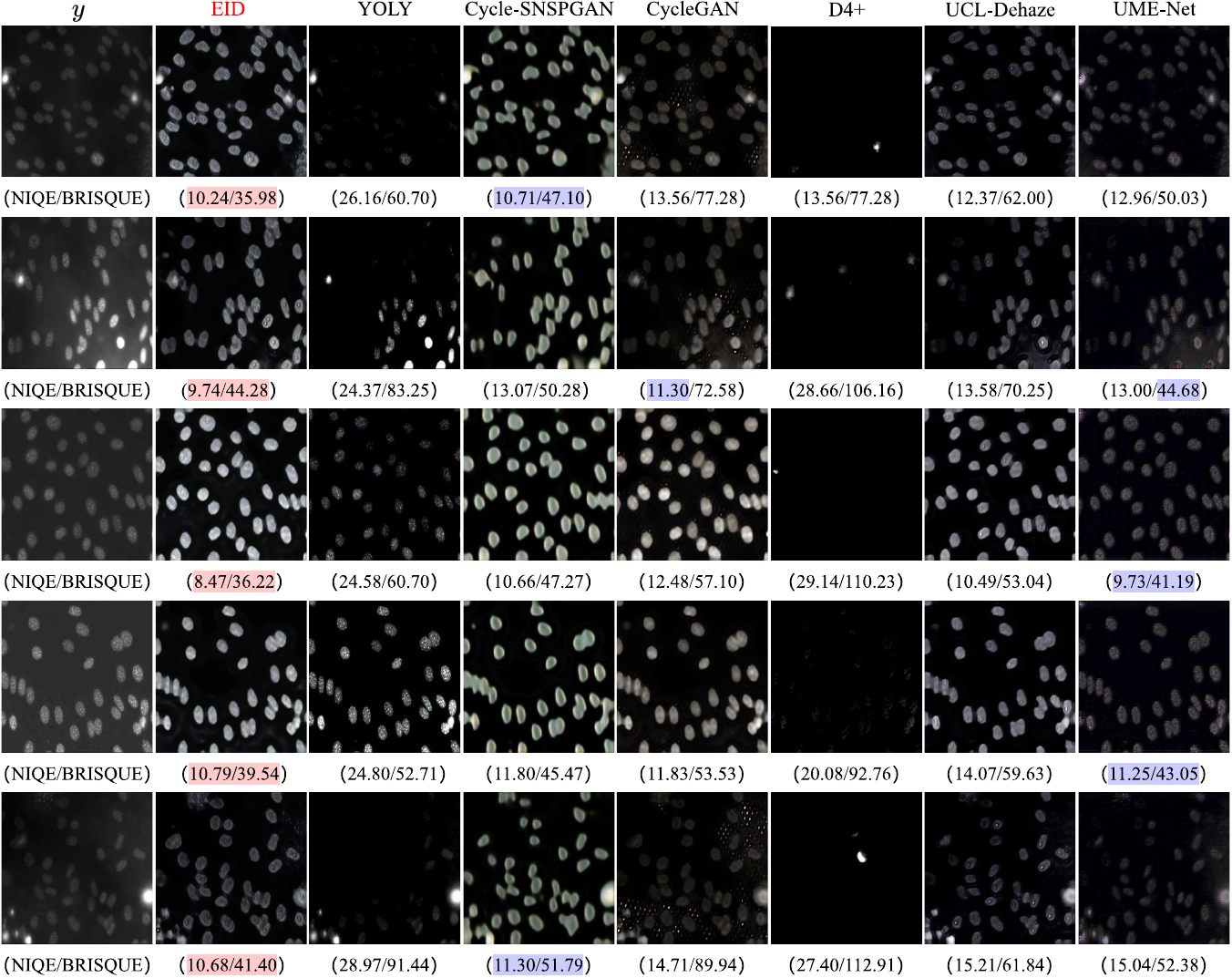}
    \caption{Results on Cell97 Dataset. Corresponding NIQE$\downarrow$ and BRISQUE$\downarrow$ are presented under the pictures. The best is \highlight{red}{red}, and the second-best is \highlight{blue}{blue}.}
    \label{kaggle-cell-result}
\end{figure}

The quantitative comparisons are presented in  Tab.~\ref{quantitative} and Figure~\ref{kaggle-cell-result}. We have the following observations. First, EID recovers higher brightness and better overall contrast, making nuclear details more visible. YOLY and CycleGAN outputs appear dimmer, losing fine details. Cycle-SNSPGAN slightly improves upon CycleGAN but suffers from color inconsistencies. D4+ performs poorly, producing overly dark images. UCL-Dehaze and UME-Net yield more stable outputs, though still below EID in terms of clarity and brightness. 

\begin{table*}[!htbp]
\centering
\renewcommand{\arraystretch}{1.5}
\resizebox{\textwidth}{!}{
\begin{tabular}{l|ccc|ccc|cc|cc}
\toprule
\multirow{3}{*}{\textbf{Method}} & \multicolumn{3}{c|}{\textbf{Medical Endoscope }} & \multicolumn{3}{c|}{\textbf{Cell Microscope}} & \multicolumn{4}{c}{\textbf{Natural Image}} \\
 &\multicolumn{3}{c|}{\textbf{Cholec80-Haze Dataset} \cite{cholec80data}}&\multicolumn{3}{c|}{\textbf{Cell 97 Dataset \cite{kagglecell}}}&\multicolumn{2}{c|}{\textbf{RESIDE-OTS \cite{RESIDE-DATASET}}} &\multicolumn{2}{c}{\textbf{RESIDE-HSTS \cite{RESIDE-DATASET}}}\\ 
 \cmidrule{2-11}
 & \textbf{NIQE}$\downarrow$& \textbf{BRISQUE}$\downarrow$& \textbf{FID}$\downarrow$& \textbf{NIQE}$\downarrow$& \textbf{BRISQUE}$\downarrow$& \textbf{FID}$\downarrow$& \textbf{PSNR}$\uparrow$& \textbf{SSIM}$\uparrow$& \textbf{PSNR}$\uparrow$& \textbf{SSIM}$\uparrow$\\

\midrule
DCP \cite{he-DCP-2010-intro} & 24.77 & 108.13 & 313.638 & \ding{56} & \ding{56} & \ding{56} & 16.29 & 0.664 & 15.50 & 0.817 \\
NLP  \cite{berman-ncp-prior2016-intro} & 21.33 & 92.99 & 286.201 & \ding{56} & \ding{56} & \ding{56} & 16.42 & 0.659 & 18.50 & 0.790 \\
YOLY \cite{yoly-baseline} & 4.49 & 27.21 & 62.773 & 22.69 & 77.85 & 436.164 & 23.89 & 0.903 & 22.90 & 0.867 \\
\midrule
Cycle-Dehaze \cite{cycledehze-intro}  & 7.96 & 25.37 & 153.267 & 17.30 & 72.43 & 515.348 & 19.95 & 0.788 & 20.95 & 0.807  \\
Cycle-SNSPGAN \cite{wang2022cycle-intro}  & 5.01 & 22.54 & 123.897 & 12.12 & 61.84 & 426.224 & 24.19 &  0.902 &  21.13 & 0.874 \\
CycleGAN \cite{cycleGAN-intro}   & 5.76 & \highlight{blue}{14.91} & 75.340 & 13.12 & 72.63 & 414.804 & 22.04  & 0.842 & 22.07  & 0.881   \\
D4+ \cite{d4+-baseline} & \highlight{blue}{4.32} & 20.73 & \highlight{blue}{61.904} & 24.63 & 100.35 & 467.949 & 24.63 & 0.901 & 22.46 & 0.874 \\
UCL-Dehaze \cite{ucldehaze-intro}  & 6.23 & 18.57 & 67.560 & 12.35 & 67.13 & \highlight{blue}{393.661} & 24.88 & 0.912 &  \highlight{blue}{22.96} &  \highlight{blue}{0.901}  \\
UME-Net \cite{umenet-2024-pr-baseline} & 5.73 & 16.94 & 76.481 & \highlight{blue}{11.03} & \highlight{blue}{50.70} & 442.052 & \highlight{blue}{25.00} & \highlight{blue}{0.915} & 22.42 & 0.891\\
\midrule
\textbf{EID}   & \highlight{red}{3.10} & \highlight{red}{4.06} & \highlight{red}{57.686} & \highlight{red}{10.68} & \highlight{red}{44.12} & \highlight{red}{387.740} & \highlight{red}{25.18} &  \highlight{red}{0.919} &  \highlight{red}{24.15} &  \highlight{red}{0.921}   \\
\bottomrule
\end{tabular}}
\caption{Performance comparison on Cell97, Cholec80-Haze, RESIDE-OTS and RESIDE-HSTS datasets. NIQE$\downarrow$, BRISQUE$\downarrow$ and FID$\downarrow$ are reported for Cell97 and Cholec80-Haze, PSNR$\uparrow$ and SSIM$\uparrow$ are reported for RESIDE-OTS and RESIDE-HSTS. The \highlight{red}{red} and \highlight{blue}{blue} markers represent the best and second best values respectively.}
\label{quantitative}
\end{table*}

\subsection{Natural Image Dehazing}

Previous results have demonstrated the effectiveness of EID in scientific imaging tasks. We wondered whether EID would achieve similar results when dehazing natural images, for which both the haze model and the ground truth images are unavailable. To this end, we perform EID on two natural image benchmarks: RESIDE-OTS \cite{RESIDE-DATASET} and  RESIDE-HSTS \cite{RESIDE-DATASET}.

\begin{figure}[!htbp]
    \centering
    \includegraphics[width=0.99\linewidth]{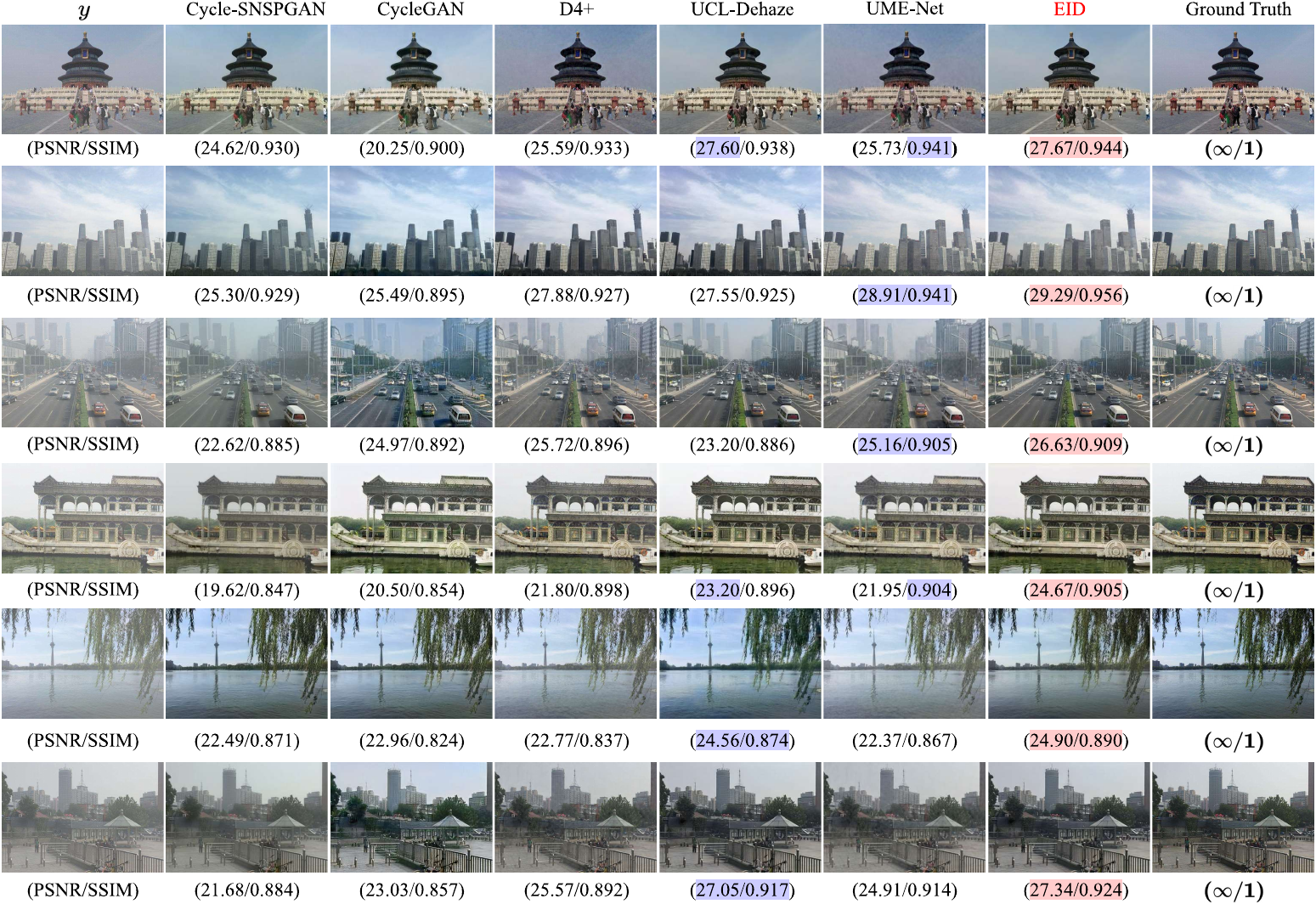}
    \caption{Results on RESIDE-OTS. Corresponding PSNR$\uparrow$ and SSIM$\uparrow$ are presented under images. The best is \highlight{red}{red}, and the second-best is\highlight{blue}{blue}.}
    \label{ots-result}
\end{figure}

In particular, we divide RESIDE-OTS into two subsets: the first is to learn the pseudo $\mathcal{H}$ which contains clear images numbered from 0 to 4499 and hazy images numbered from 4500 to 8700. The second is to train EID, only the hazy images from 4500 to 8700 are used. Finally, the remaining hazy images are the test dataset. In order to comprehensively evaluate the dehazing performance of EID, we also perform test experiments on the RESIDE-HSTS dataset. The first two datasets are not ground truth datasets at all. The results on these two datasets can prove that our method can remove the blur on images without ground truth. The last two datasets have the ground truth of the hazy images. 

The comparisons are shown in Figure~\ref{ots-result}. As can be seen, the EID model outperforms all other methods in terms of dehazing effectiveness, color restoration, detail preservation, and contrast. UME-Net is the second best model, although it has slightly lower brightness and suboptimal detail recovery. D4+ shows strong dehazing capabilities, but its overall brightness and color restoration remain inferior to EID. UCL-Dehaze shows relatively good color fidelity in certain scenarios, but suffers from a lack of detail and contrast. Cycle-SNSPGAN and CycleGAN perform poorly in dehazing tasks, with incomplete haze removal and significant loss of detail. 

\subsection{Ablation Study}

\begin{figure}[!htbp]
    \centering
\begin{subfigure}{\textwidth}
        \centering
        \includegraphics[width=\textwidth]{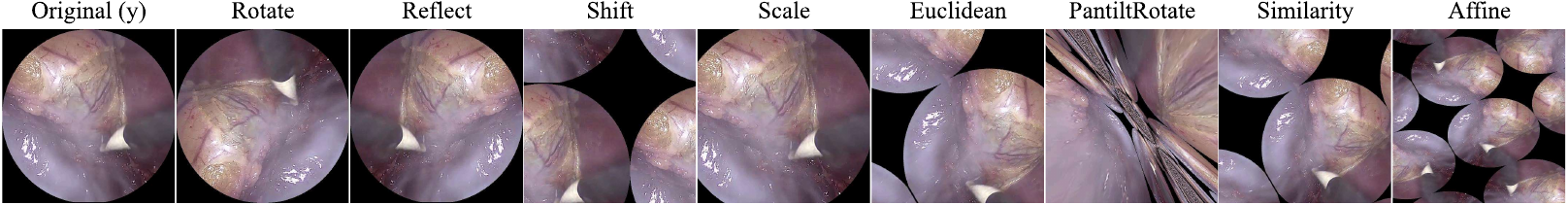}
        \caption{Examples of different transformations on endoscope data.}
    \end{subfigure}
    
    \vskip\baselineskip 
    
    \begin{subfigure}{0.48\linewidth}
        \centering
        \includegraphics[width=\textwidth]{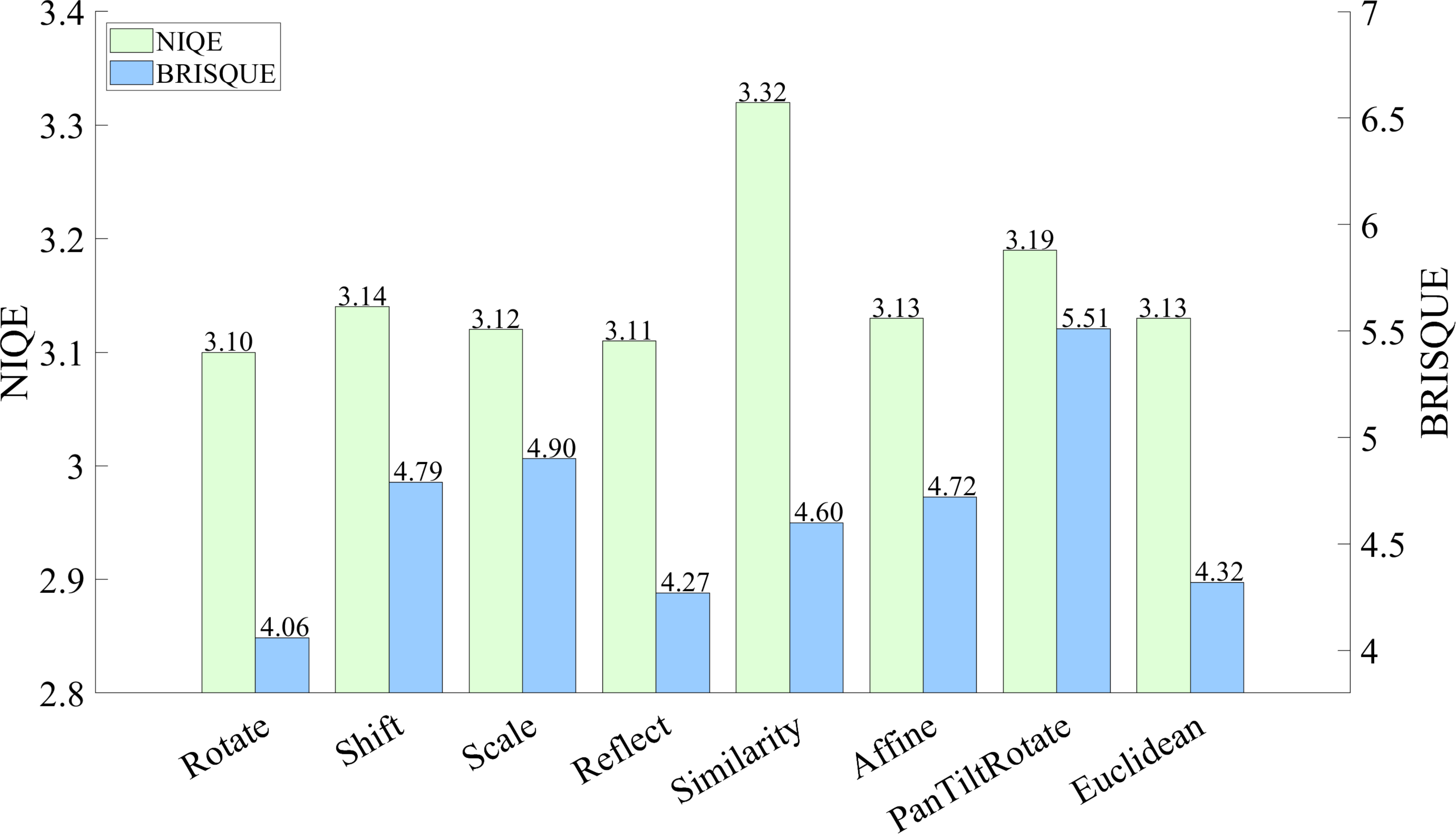} 
        \caption{Endoscope (Cholec80-Haze). NIQE$\downarrow$ and BRISQUE$\downarrow$ are shown.}
        \label{transformation}
    \end{subfigure}
    \hfill
    \begin{subfigure}{0.48\linewidth}
        \centering
        \includegraphics[width=\textwidth]{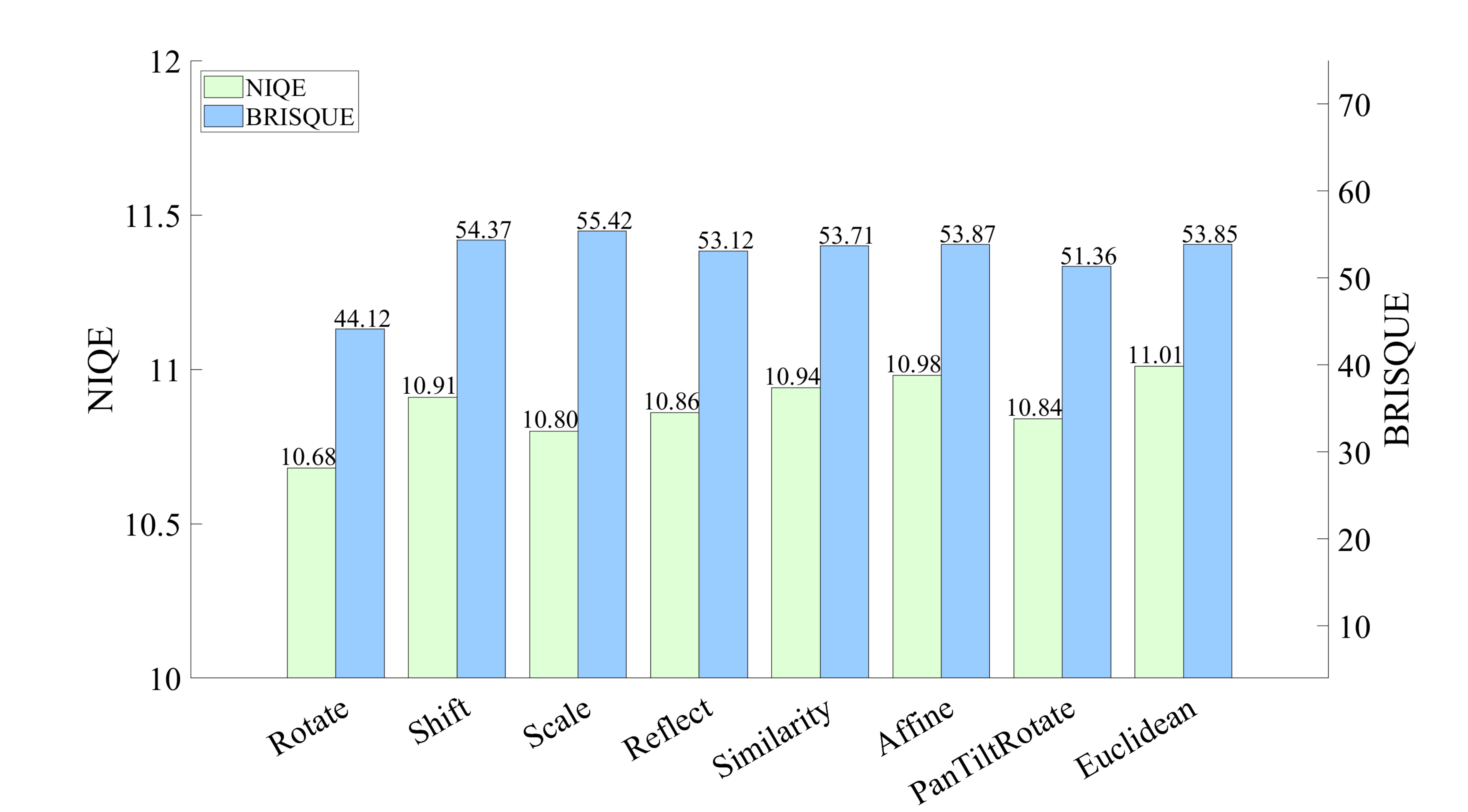} 
        \caption{Cell microscopy (Cell97). NIQE$\downarrow$ and BRISQUE$\downarrow$ are shown.}
    \end{subfigure}

    \vskip\baselineskip  

    \begin{subfigure}{0.48\linewidth}
        \centering
        \includegraphics[width=\textwidth]{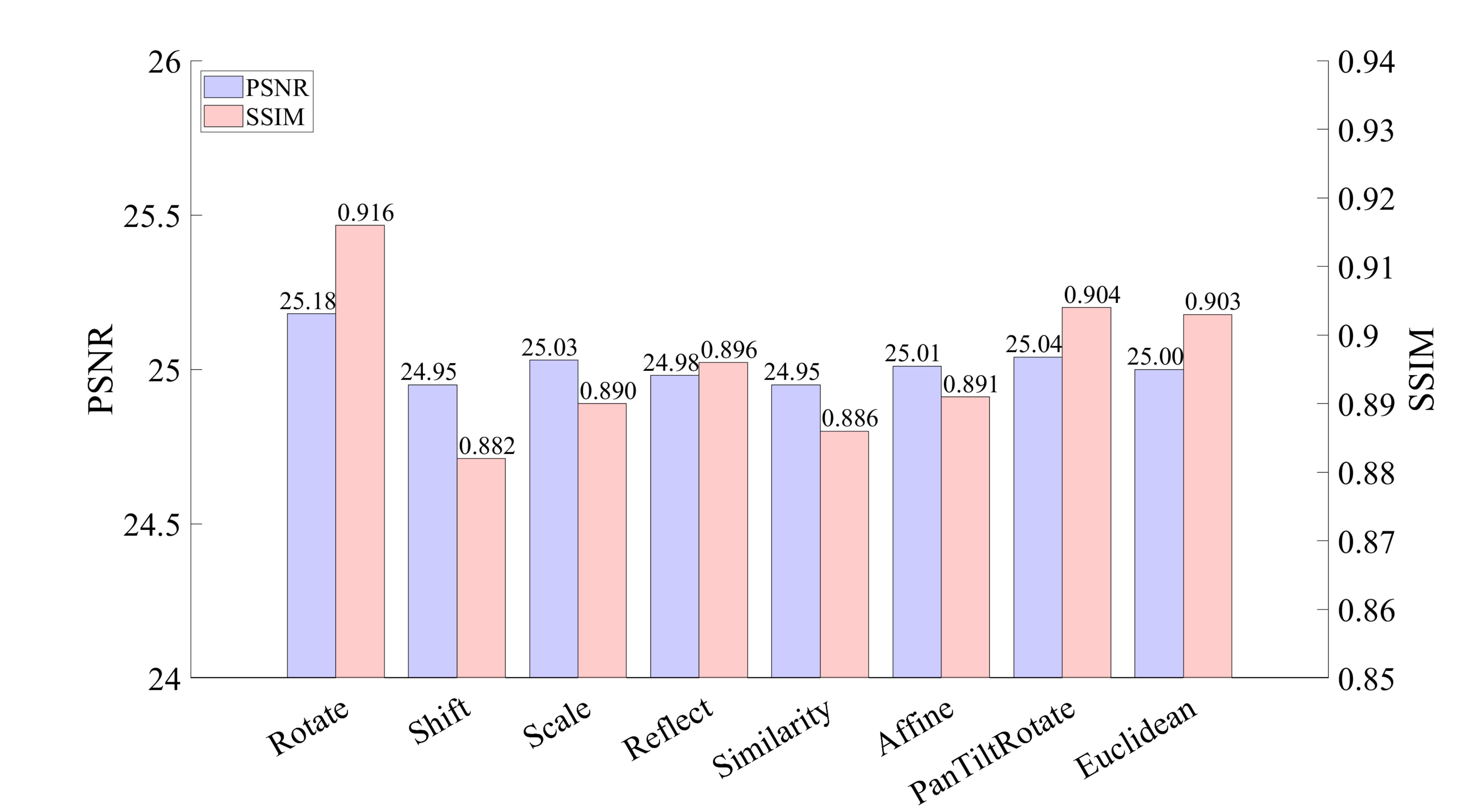}
        \caption{Natural images (RESIDE-OTS). PSNR$\uparrow$ and SSIM$\uparrow$ are shown.}
    \end{subfigure}
    \hfill
        \begin{subfigure}{0.48\linewidth}
        \centering
        \includegraphics[width=\textwidth]{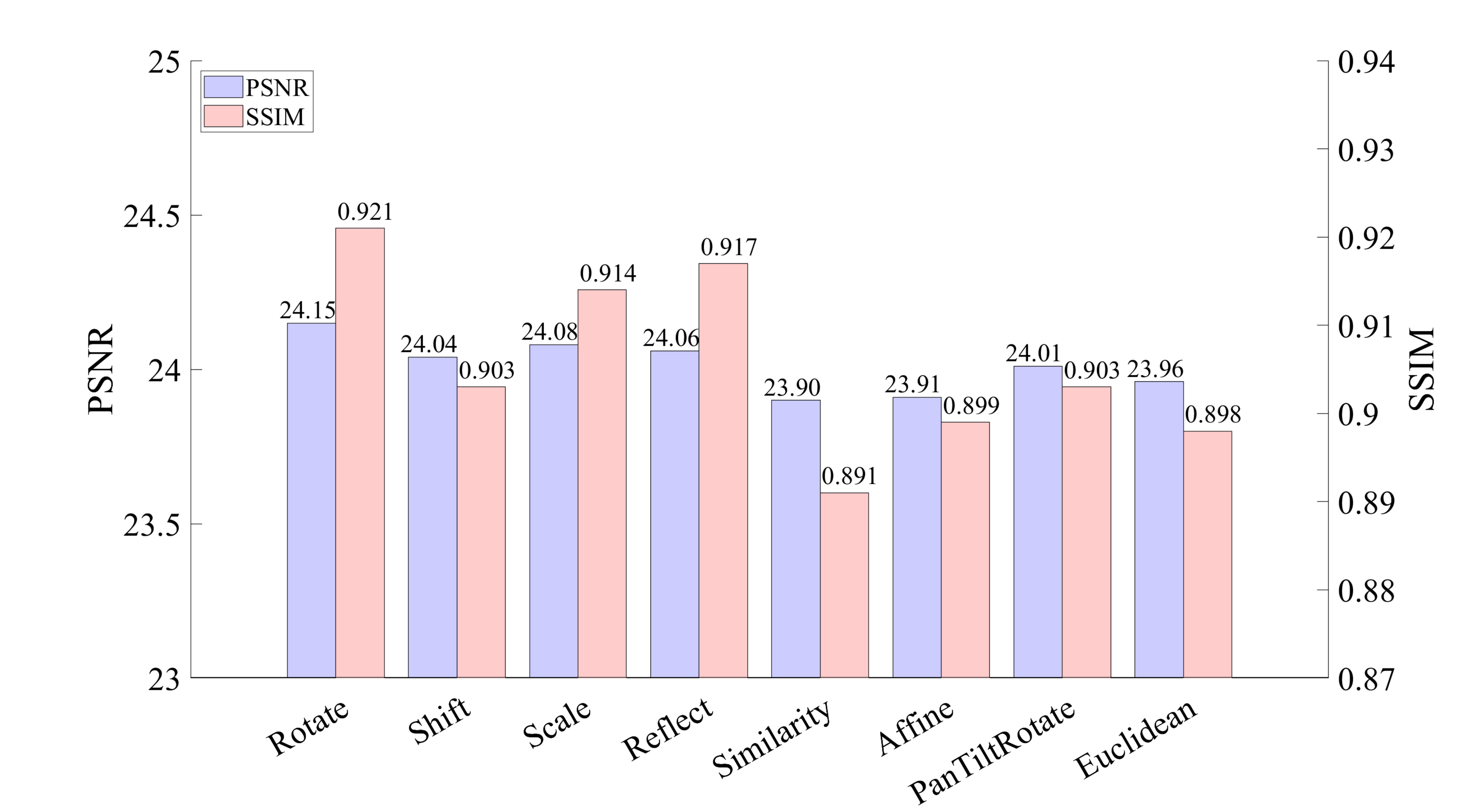} 
        \caption{Natural images (RESIDE-HSTS). PSNR$\uparrow$ and SSIM$\uparrow$ are shown.}
    \end{subfigure}
    

    \caption{Ablation study on the transformations in EID. (a) Examples of different transformations. (b)-(e) Performance of EID using different transformations on different applications. }
    \label{fig:transform}
\end{figure}

\noindent\textbf{On the impact of transformation.} Selecting an appropriate transformation group is crucial for equivariant learning without ground truth \cite{dongdongEI2021-intro,spm}. In this section, we explore the effects of eight popular transformations that are well-defined in \cite{deepinv}, including four basic transformations: Rotate, Shift, Scale, and Reflect, and four projective transformations: Similarity, Affine, PanTiltRotate and Euclidean, as shown in Figure\ref{fig:transform} (a). The results for Cholect80-Haze, Cell97, RESIDE-OTS and RESIDE-HSTS are shown in Figure\ref{fig:transform} (b-e). We note that, among all the transformations, rotation achieves the best performance on both datasets. Furthermore, the small differences between the transformations suggest that clean images exhibit invariance to a wider range of transformations. Finally, we also conducted experiments on mixed transformations, the results of which are shown in Tab.\ref{transform_sm}. The results indicate that none of the mixed transformations outperform the rotation transformation. This implies that the datasets are mainly dominated by rotation invariance, which also motivates the choice of rotation transformations for all subsequent experiments.

\begin{table*}[t]
\centering
\renewcommand{\arraystretch}{1.3}
\resizebox{\textwidth}{!}{
\begin{tabular}{l|cc|cc|cc|cc}
\toprule
\multirow{2}{*}{\textbf{Method}} & \multicolumn{2}{c|}{\textbf{Cholec80-Haze}} & \multicolumn{2}{c|}{\textbf{Cell97}} & \multicolumn{2}{c|}{\textbf{OTS outdoor}} & \multicolumn{2}{c}{\textbf{HSTS}} \\
& \textbf{NIQE}$\downarrow$ & \textbf{BRISQUE}$\downarrow$ & \textbf{NIQE}$\downarrow$ & \textbf{BRISQUE}$\downarrow$ & \textbf{PSNR}$\uparrow$ & \textbf{SSIM}$\uparrow$ & \textbf{PSNR}$\uparrow$ & \textbf{SSIM}$\uparrow$ \\
\midrule

Rotate  & \highlight{red}{3.10} & \highlight{red}{4.06} & \highlight{red}{10.68} & \highlight{red}{44.12} & \highlight{red}{25.18} & \highlight{red}{0.919} & \highlight{red}{24.15} & \highlight{red}{0.921}  \\

\midrule

Rotate+Shift & \highlight{blue}{3.18} & 6.17 & 10.80 & \highlight{blue}{44.85} & 24.56 & 0.883 & 22.29 & 0.851 \\

Rotate+Affine & 3.27 & 5.57 & 10.82 & 46.08 & 24.32 & 0.879 & 22.46 &  0.847 \\

Rotate+Euclidean & 3.27 & 4.77 & 10.76 & 45.05 & 24.51 & 0.887 & 22.81 & 0.858 \\

Rotate+PanTiltRotate & 3.30 & 5.84 & \highlight{blue}{10.73} & 46.41 & \highlight{blue}{24.78} & \highlight{blue}{0.890} & \highlight{blue}{22.94} & 0.862 \\

Rotate+Similarity & 3.26 & 4.88 & 10.76 & 46.32 & 23.66 & 0.877 &  22.55 & 0.851 \\

Shift+PanTiltRotate & 3.28 & 4.74 & 10.91 & 46.11 & 23.46  & 0.863 & 21.91  & 0.830   \\ 

\midrule

PanTiltRotate+Affine & 3.24 & 4.65 & 10.89 & 45.66 & 24.05 & 0.878 & 22.52 &  0.849 \\

PanTiltRotate+Euclidean & 3.26 & \highlight{blue}{4.47} & 10.86 & 46.182 & 23.95 & 0.866 & 22.24 & 0.837 \\

PanTiltRotate+Similarity & 3.25 & 4.70 & 10.91 & 44.89 & 24.08 &  0.879 & 22.75 & \highlight{blue}{0.863}  \\ 

\bottomrule
\end{tabular}
}
\caption{
Ablation study on transformation combinations used in EID. For each configuration, we report NIQE$\downarrow$ and BRISQUE$\downarrow$ on Cholec80-Haze and Cell97 datasets (scientific imaging), and PSNR$\uparrow$, SSIM$\uparrow$ on RESIDE-OTS and RESIDE-HSTS (natural images). \highlight{red}{Red} and \highlight{blue}{Blue} indicate best and second-best scores, respectively.
}
\label{transform_sm}
\end{table*}

\noindent\textbf{On the impact of loss components.}
We trained three EID variants, each using a different combination of the loss functions, $\mathcal{L}_{hc}$ and$\mathcal{L}_{ec}$. V1 uses only the former, V2 uses only the latter and V3 uses both.  The results on the Cholec80-Haze and RESIDE-OTS datasets are shown in Tab.~\ref{ablation-combined}, demonstrating that: (1) Using only one loss function (e.g. V1 or V2) leads to suboptimal results. (2) Integrating both loss functions enables EID to achieve optimal results. This is because the equivariance loss $\mathcal{L}_{ec}$ mainly focuses on recovering clean patterns in the nullspace of $\mathcal{H}$, while restricting learning outside the range space of $\mathcal{H}$ whose patterns must be learned using $\mathcal{L}_{hc}$. This further demonstrates our motivation for jointly learning with physics-informed and equivariant self-supervision to recover clean images.

Finally, although all the results are encouraging, we believe there is still room to improve EID's performance further by integrating state-of-the-art network architectures \cite{EMMA,zhao2024EIfusion1-relate,pr2025eenet}. We will leave this as our future work.

\begin{table}
\centering
\small
\resizebox{0.94\linewidth}{!}{ 
\begin{minipage}{0.58\linewidth}
    \centering
    \begin{tabular}{lccc}
    \toprule
    loss  & \textit{V$_1$} & \textit{V$_2$} & \textit{V$_3$}  \\
    \midrule
    $\mathcal{L}_{hc}$   & \ding{51} & w/o & \ding{51}\\
    $\mathcal{L}_{ed}$  & w/o & \ding{51} & \ding{51}\\
    \midrule
     NIQE$\downarrow$& 5.54 & 6.08 & 3.10  \\
     BRISQUE$\downarrow$ & 8.46 & 10.17 & 4.06\\
    \bottomrule
    \end{tabular}
\end{minipage}
\hfill
\begin{minipage}{0.58\linewidth}
    \centering
    \begin{tabular}{lccc}
    \toprule
    loss  & \textit{V$_1$} & \textit{V$_2$} & \textit{V$_3$}  \\
    \midrule
    $\mathcal{L}_{hc}$   & \ding{51} & w/o & \ding{51}\\
    $\mathcal{L}_{ed}$  & w/o & \ding{51} & \ding{51}\\
    \midrule
     PSNR$\uparrow$ & 24.27 &  19.57  & 25.18 \\
     SSIM$\uparrow$ & 0.902 & 0.804 & 0.919\\
    \bottomrule
    \end{tabular}
\end{minipage}}
\caption{Ablation study on the different components in loss functions. Performance of EID using different loss function on Cholec80-Haze is showed on left table, and the right table shows EID performance on RESIDE-OTS.}
\label{ablation-combined}
\end{table}

\section{Discussion}

Despite its promising performance, our study has several limitations. Firstly, the equivariant imaging properties of images differ between various applications. For instance, the strict symmetry properties exhibited by natural images may not always hold in certain complex scenes, which could affect performance in highly structured or non-homogeneous environments. Secondly, although our proposed pseudo-physics estimation technique enables us to learn a haze model unsupervisedly, its accuracy still depends on the quality of adversarial training and the representation capacity of neural networks. Thirdly, our framework relies on computationally intensive deep learning models, which may pose challenges for real-time deployment in environments with limited resources. 

There are several improvements and extensions that could be explored in future work. One potential approach is to increase the adaptability of EID by incorporating domain-specific priors for various imaging tasks. Improving the efficiency of our model through lightweight architectures or knowledge distillation could enable real-time image dehazing on embedded systems. Another avenue for exploration is integrating multimodal data, such as textual information or infrared imagery \cite{zhao2024EIfusion1-relate}, to further refine the dehazing process in complex environments. Furthermore, EID’s performance could be improved by using supervised learning methods to model haze physics \cite{EMMA}. Finally, our framework could be adapted to solve other inverse imaging problems besides image dehazing, such as underwater image enhancement and remote sensing image restoration.

\section{Conclusion}
In this paper, we presented EID, a novel unsupervised learning paradigm for image dehazing. EID leverages the inherent symmetry properties of scientific and natural images to achieve high-quality dehazing without requiring ground truth supervision. Extensive experimental results demonstrated that EID achieves state-of-the-art dehazing performance on medical endoscopy, cell microscopy, and natural images. By bridging the gap between physics-based modelling and data-driven learning, we hope our approach will shed new light on unsupervised image dehazing and broader scientific imaging scenarios.

{
\bibliographystyle{unsrt}
\bibliography{reference}
}

\end{document}